\documentclass[twocolumn]{article}
\usepackage{spconf,amsmath,graphicx, lipsum}
\usepackage{dsfont}
\usepackage{caption} 
\usepackage{multirow}
\usepackage{graphicx}

\title{Robustness of Saak transform against Adversarial Attacks}

 \name
{Thiyagarajan Ramanathan$\dagger$, Abinaya Manimaran$\dagger$, Suya You$\ddagger$, C-C Jay Kuo$\dagger$}
\address	
{$\dagger$University of Southern California, Los Angeles, California, USA \\
 $\ddagger$US Army Research Laboratory, Playa Vista, California, USA}

\begin{document}

\maketitle

\begin{abstract}

Image classification is vulnerable to adversarial attacks. This work investigates the robustness of Saak transform against adversarial attacks towards high performance image classification. We develop a complete image classification system based on multi-stage Saak transform. In the Saak transform domain, clean and adversarial images demonstrate different distributions at different spectral dimensions. Selection of the spectral dimensions at every stage can be viewed as an automatic denoising process. Motivated by this observation, we carefully design strategies of feature extraction, representation and classification that increase adversarial robustness. The performances with well-known datasets and attacks are demonstrated by extensive experimental evaluations. 

\end{abstract}

\begin{keywords}
Saak transform, Adversarial attacks, Deep Neural Networks, Image Classification
\end{keywords}

\section{Introduction} \label{sec:Introduction}

It has been shown that deep learning based approach for image classification is vulnerable to adversarial attacks \cite{BIM}. These attacks come in the form of adversarial inputs with carefully crafted perturbations added to the input samples. Such perturbations are small and imperceptible to humans, but can drastically cause the classification systems to misinterpret adversarial inputs, with potentially disastrous consequences where safety and security are crucial.

Saak (Subspace approximation with augmented kernels) transform, inspired by deep learning mechanism, is a new mathematical transform that is completely interpretable \cite{DBLP:journals/corr/Kuo16, DBLP:journals/corr/Kuo17}. The Saak transform is a variant of principal component analysis (PCA) that splits the positive and negative responses into two separate channels by kernel augmentation and resolves "sign confusion" ambiguity. This process facilitates the cascade of Saak transforms called multi-stage Saak transforms. Saak features at later stages have larger receptive fields, yet they are obtained in a one-pass feed-forward manner without any supervision and back propagation. In addition, inverse of Saak transform is possible, which allows the Saak feature representations to be transformed back to the image space for clearly visualizing, analyzing, and interpreting.

Saak transform has demonstrated its superior performance and utility in classifying hand-written digits under various noisy environments \cite{DBLP:journals/corr/abs-1710-10714}. Furthermore, Saak transform has also been employed as a pre-processing step in image classification pipeline to defend adversarial attack \cite{song2018defense}. 

In this work, we investigate the robustness of Saak transform against adversarial attacks towards high performance image classification. We develop a complete image classification system based on multi-stage Saak transform. We take advantage of the ocean of Saak coefficients available at every stage of multi-stage Saak transform. Careful selection of these features using cross-entropy leads us build a new Saak feature representation. The whole feature extraction and selection process is completely transparent and of extremely low complexity. In the Saak transform domain, clean and adversarial images demonstrate different distributions at different spectral dimensions. Selection of the spectral dimensions at every stage can be viewed as an automatic denoising process. Motivated by this observation, we design new strategies of feature extraction, representation and classification that increase adversarial robustness. The performances with well-known benchmark datasets and attacks are demonstrated by extensive experimental results. 

\section{Related Work} \label{sec:Related Work}
 
One of the most interesting explorations in defense against adversarial attacks is done through adversarial training \cite{Adversarial}. This aims in augmenting adversarial samples along with clean samples for simultaneous training. While these methods help in defending against particular adversarial attacks for which it is trained for, they fail to generalize. Also, this type of training takes longer time for convergence, hence needs to be trained for more epochs. 

Adversarial detection involves detection of an adversarial sample before passing through the network. Adversarial samples can be detected using statistical tests \cite{DBLP:journals/corr/GrosseMP0M17}, estimating Bayesian uncertainty \cite{Feinman2017DetectingAS}, using noise reduction methods like scalar quantization and spatial smoothing filter \cite{DBLP:journals/corr/LiangLSLSW17}. Though these methods pave way to a good adversarial sample detection problem, these detectors still possess the risk of being fooled by the attacker. 

Pre-processing based methods perform certain transformations on inputs to nullify the effect of adversarial attacks. Some examples are image cropping and rescaling, JPEG compression \cite{DBLP:journals/corr/DasSCHCKC17}, feature squeezing by Bit-Depth-Reduction \cite{DBLP:journals/corr/XuEQ17}, and Total Variance Minimization \cite{DBLP:journals/corr/abs-1711-00117}. Nonlinear, saturating neural networks are used in \cite{DBLP:journals/corr/NayebiG17}. Gradient masking effect applies regularizers or smooth labels to attain output less sensitive to perturbed input \cite{DBLP:journals/corr/PapernotMSW16}.

Use of knowledge distillation when training networks can be used as defense against adversarial samples \cite{DBLP:journals/corr/PapernotMWJS15}. Reinforcement of network structure by using bounded ReLU activations help in enhancing stability to adversarial perturbations \cite{DBLP:journals/corr/ZantedeschiNR17}. Pixel defend is used as an image purification process, as described in \cite{DBLP:journals/corr/abs-1710-10766}.

\cite{song2018defense} applies lossy Saak transform to adversarially perturbed images as a pre-processing tool to defend against adversarial attacks. The method is based on the observation that outputs of Saak transform are very discriminative in differentiating adversarial examples from clean ones. Instead of using Saak transform as pre-processing tool, we apply multi-stage Saak transform to build a complete image classification pipeline and design new strategies of feature selection, representation and classification to defend against adversarial attacks.

\section{SAAK Transform Approach} \label{sec:SAAK Transform Approach}

The Saak transform defines a mapping from a real-valued function defined on a three-dimensional cuboid consisting of spatial and spectral dimensions to a one-dimensional rectified spectral vector. It is presented as a new feature representation method. It consists of two main ideas: subspace approximation and kernel augmentation. For the  former, we build the optimal linear subspace approximation to the original signal space via PCA or the truncated Karhunen-Loève Transform (KLT) \cite{Stark:1986:PRP:5639}. For the latter, we augment each transform kernel with its negative and apply the rectified linear unit (ReLU) to the transform output. This is equivalent to the sign-to-position (S/P) format conversion. 

\begin{figure}
\begin{center}
\includegraphics[width=1\linewidth]{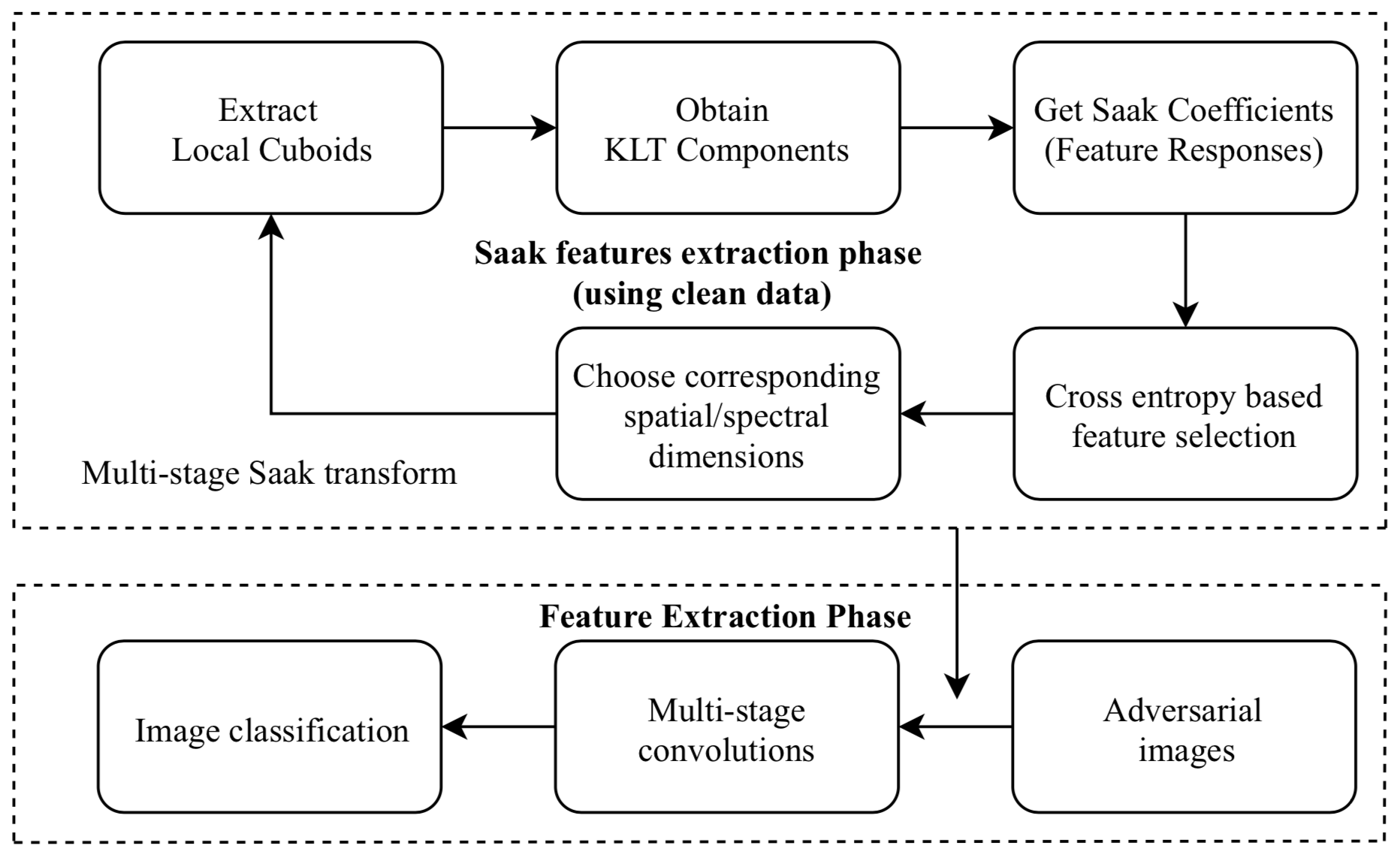}
\end{center}
\vspace*{-5mm}
\caption{Saak transform consists two modules: kernel extraction and feature extraction. We use clean training images to extract kernels followed feature extraction. Feature responses can used for any application.}
\label{fig:flow chart}
\end{figure}

Figure \ref{fig:flow chart} illustrates the developed saak transform pipeline. Specifically, it consists of 1) Extracting Local Cuboids from the images, 2) Obtaining KLT components, 3) Convoluting the images with the extracted kernels, 4) Calculating the cross entropy measures, 5) Selecting the best spatial/spectral components. We classify adversarial images using Saak transform. We extract kernels using clean images and follow the same procedure as we classify clean images. Saak kernels are used to extract the coefficients from attacked images. We classify adversarial attacked images after selecting features using our cross-entropy based method.

During the classification phase, input \textbf{f} is convoluted with extracted Saak kernels to extract Saak features. Based on kernel size $k_{s}$, spatial resolution of feature 
responses reduce at every stage. Consider block of feature responses at any stage $p$ for a single image as {\boldmath${f_{p}}$} with dimension $D_{p1} \times D_{p2} \times K_p$. First two dimensions represent spatial dimension along vertical and horizontal directions. The third one represents the spectral dimension of feature responses for an image. If there are $N$ images in the training data, then total dimension of feature responses can be given as $N \times D_{p1} \times D_{p2} \times K_p$. 
Cross-entropy for feature responses is calculated at every index $(i,j,k)$, where $(i,j)$ represents spatial location and $k$ represents spectral dimension. Let $C$ be the number of classes. Entropy at every location is given by,

\begin{equation}\label{eqn:cross entropy}
    H = \sum_{n=1}^{N}\sum_{c=1}^{C}{y_{n,c}\hspace{0.5mm}\log{\frac{1}{p_{n,c}}}}
\end{equation}
where
\begin{equation}\label{eqn:ync}
   y_{n,c} = 
   \begin{cases}
            1, &  \text{if } \boldsymbol{f_p}(n,i,j,k) \in c\\
            0, &  \text{if } \boldsymbol{f_p}(n,i,j,k) \notin c
    \end{cases}
\end{equation}

and $p_{n,c}$ is the probability of $n^{th}$ sample in class $c$. To obtain this, feature response values at $(i,j,k)$ location across all images are taken. Histogram of these $N$ values is calculated using a certain number of bins, $B$. From various experiments, we concluded that feature selection is stable irrespective of number of 
bins. We choose $B =10$ and proceed by getting

\begin{equation}\label{eqn:bins entropy}
    mc = (mc_1, mc_2, ..., mc_B),
\end{equation}
where $mc_i$ represents maximum occurring class in bin $i$, and $mc_i
\in {1,2,...,C}$. Probability $p_{n,c}$ is determined as
\begin{equation}\label{eqn:pnc}
    p_{n,c} = \frac{\sum_{i=1}^{B}\mathds{1}(mc_i = c)}{B}
\end{equation}

At the end, $D_{p1} \times D_{p2} \times K_{p}$ cross-entropy values will be computed at stage $p$. Lower the entropy value at a location, higher is the discriminant power. For every spectral dimension, $D_{p1} \times D_{p2}$ pixels are ranked according to their entropy. The first few pixels with lowest cross-entropy values are retained, and 
others are made zero. This localizes salient regions in an image. Similarly, spatially averaged cross-entropy for all spectral dimensions is obtained. From these average values, spectral dimensions are ranked, and first few $K_p^{'}$ with lowest average cross-entropy values are chosen. Thus spatially sparse feature responses with dimension 
$D_{p1} \times D_{p2} \times K_p^{'}$ are chosen at stage p. This is repeated at all stages of multi-stage Saak transform for classification.

\section{Adversarial Attacks and Defenses} \label{sec:Adversarial Attacks and Defenses}

We consider three major adversarial attacks, against which we will evaluate our approach. The attacks are explained below. 

\textbf {Fast Gradient Sign Method (FGSM) \cite{goodfellow2014explaining}}: This method computes adversarial image by adding a pixel-wide perturbation of magnitude in the direction of gradient. The perturbation is computed by $\eta = \epsilon sign (\Delta_x J_\theta(x,l))$,  so each pixel is modified by $x'=x+\eta$. This value can be computed using back propagation. There is no bound on the modified value, hence the quality of the adversarial image greatly decreases.

\textbf{Basic Iterative Method (BIM) \cite{DBLP:journals/corr/KurakinGB16a}}: This method is an extension of FGSM, but with a limit on the value that a pixel can be modified. The change is limited, but the number of iterations of the attack are increased. Hence, for human eye, the BIM attacked images look less noisier when compared to FGSM attacks. The adversarial images are generated after multiple iterations. 

\textbf{DeepFool (DF) \cite{DBLP:conf/cvpr/Moosavi-Dezfooli16}}: This attack is more carefully crafted when compared to FGSM and BIM. It computes the closest $L2$ projection distance to the decision boundary hyper-plane of adversarial example and input image. The perturbation is a function of the $argmin$ of this distance. The perturbation is applied iteratively with smaller steps, hence the produced adversarial images do not look noisy to human eye.  

The above three attacks are used to generate adversarial images and our Saak transform based image classification are directly applied to classify the attack images. The classification accuracy is evaluated against following state-of-the-art defensing techniques.  

\textbf{JPEG Compression \cite{DBLP:journals/corr/DasSCHCKC17}}: 
It has demonstrated that systematic JPEG compression can work as an effective pre-processing step in the classification pipeline to counter adversarial attacks. An important component of JPEG compression is its ability to remove high-frequency signal components, hence reducing high-frequency components with JPEG compression should contribute to adversarial attack removal without hurting the decision accuracy of clean data. 

\textbf{Bit Depth Reduction \cite{DBLP:journals/corr/abs-1711-00117}}:
It is one of the feature squeezing techniques. Images often contain unnecessary features that can be exploited by adversarial attacks. Using less bit to discrete colors will make prediction more robust. The more bit to reduce, the more features are eliminated. In this work, two variants of bit depth reduction (4-bits and 5-bits) are used to pre-process the attack image.

\textbf{Non-Local Means (NL-Means) \cite{DBLP:journals/corr/abs-1812-03411}}: Image denoising using NL-means before classification can remove adversarial noise and improve adversarial robustness. The NL-means not only compares the pixel value in a single point but the geometrical configuration in a whole neighborhood. This fact allows a more robust comparison than neighborhood filters, improving robustness of networks against adversarial attacks.

\textbf{Total Variance Minimization (TVM) \cite{DBLP:journals/corr/abs-1711-00117}}:
In images, noisy signals have a high total variation. TVM is an optimization technique where the variance of the pixels are reduced using L2 regularization. TVM retains more information such as edges when compared to other filtering techniques.

\textbf{Pixel Deflection \cite{DBLP:journals/corr/abs-1801-08926}}:
A random pixel is replaced by another random pixel in a local neighborhood. It works well due to the assumption that adversarial attacks rely on specific activation functions, i.e., only some pixels are manipulated to make the attack work. There are two variations of this method - pixel deflection with and without activation map. The activation map is used to determine the random pixel which is to be used to replace the target pixel in consideration. 

\begin{figure}
\begin{center}
\includegraphics[width=1\linewidth]{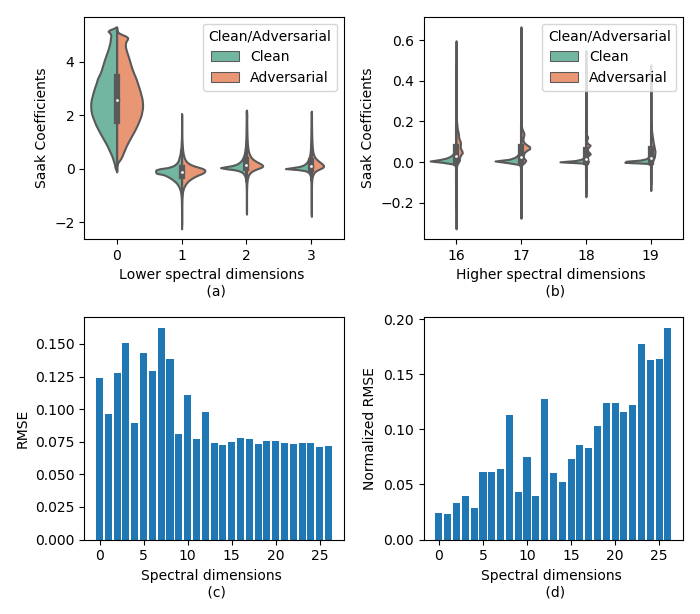}
\end{center}
\vspace*{-5mm}
\caption{(a) and (b) show the lower and higher 
spectral distribution of Saak coefficients 
extracted from CIFAR-10. At higher dimensions, the distributions obtained from clean and attacked images
are different. (c) and (d) show the RMSE and 
normalized RMSE between clean and FGSM attacked Saak coefficients in different spectral dimensions}
\label{fig:violin plot}
\end{figure}

\section{Experimental Results}\label{sec:Experimental Results}

Extensive experiments are conduced on datasets MNIST, CIFAR-10 and STL-10. We provide in-depth experimental results for adversarial images classification and prove that classification using the proposed Saak features is adversarial robust in comparison with state-of-the-art defense mechanisms. 

In the Saak transform domain, clean and adversarial images demonstrate different distributions at different spectral dimensions. Figure \ref{fig:violin plot} shows distribution of Saak components belonging to first few spectral dimensions, followed by the distribution for higher spectral dimensions. Saak spectral components differ for both clean and adversarial images at higher dimension. We also show the normalized and the original RMSE (root-mean-squared-error) values between clean and FGSM adversarial samples in different spectral components. We can observe from Figure \ref{fig:violin plot} (c) and (d) that clean and adversarial samples have different Saak coefficient values in high spectral dimensions. These results were obtained from first stage Saak transform of CIFAR-10 images using $3 \times 3$ local cuboids.

We classify adversarial images using Saak transform. We extract kernels using clean images and follow the same procedure as we classify clean images. As shown in Figure \ref{fig:flow chart} Saak kernels are used to extract the coefficients from attacked images. We classify adversarial attacked images after selecting features using cross-entropy based method.

\begin{table}[t]
\centering
\resizebox{\columnwidth}{!}{%
\begin{tabular}{ |c|c|c|c|c|c| } 
 \hline
 \textbf{Adversarial Defense} & \textbf{FGSM} & \textbf{BIM} & \textbf{DF}\\ 
 \hline
 No-defense & 13.86\%& 18.95\% &   13\% \\
 JPEG(Q=90) & 4.61\%  &9.52\% & 14.1\% \\ 
 Bit Depth Reduction (4-bit) & 5.66\% &9.65\%&  12.83\%\\
 Bit Depth Reduction (5-bit) &  3.63\% & 8.87\%&   13.06\%\\ 
 Median Filtering (2x2) & 4.04\% & 9.41\%&   8.77\%\\ 
 Median Filtering (3x3) &  5.14\% &10.2\%&  9.29\%\\
 NL-Means & 5.27\% &9.54\%&  14.52\%\\
 TVM & \textbf{2.92\%}& 8.81\%& 5.81\% \\
 Pixel Deflection (W/o RCAM) & 4.54\%& 9.55\%& 16.44\% \\
 Pixel Deflection (W/ RCAM) & 5.16\%  & 9.63\%& 17.3\%\\
 \hline
 Saak Transform & 4.78\%& \textbf{5.17\%} & \textbf{3.79\%} \\
 \hline
 \end{tabular}
 }

\caption{Robustness comparison on MNIST: The accuracy on clean images using the modified LeNet architecture is 99.2\% and using SAAK transform is 99.4\%. The value in each entry of the table is the drop in classification accuracy from clean and adversarial attacked images.}
\label{table:MNIST}
\end{table}

\begin{table}[t]
\centering
\resizebox{\columnwidth}{!}{%
\begin{tabular}{ |c|c|c|c| } 
 \hline
 \textbf{Adversarial Defense} & \textbf{FGSM} & \textbf{BIM} & \textbf{DF}\\ 
 \hline
 No-defense& 83.95\%& 83.95\%&  83.95\%\\
 JPEG(Q=90)&  74.69\%& 76.26\%&   \textbf{2.97\%} \\ 
 Bit Depth Reduction (4-bit)& 82.08\%& 82.94\%&     60.35\%\\ 
 Bit Depth Reduction (5-bit) &  81.95\%& 82.93\%&    60.8\%\\ 
 Median Filtering (2x2) & 77.87\%&  77.19\%& 71.03\%\\ 
 Median Filtering (3x3) & 82.55\%& 78.44\%&  79.99\% \\
 NL-Means & 77.41\%& 75.27\%&  3.15\%\\
 TVM & 76.18\%& 76.49\%& 72.35\%\\
 Pixel Deflection (W/o RCAM) & 83.02\%& 83.61\%&  60.06\%\\
 Pixel Deflection (W/ RCAM) &82.8\%& 83.67\% & 60.01\% \\
 \hline
 Saak Transform &\textbf{25.1\%} & \textbf{26.1\%}&  4.16\% \\
 \hline
 \end{tabular}
 }
\caption{Robustness comparison on CIFAR-10: The accuracy on clean images using pre-trained VGG-16 is 93.95\% and using SAAK transform is 74.6\%.}
\label{table:CIFAR-10}
\end{table}

\begin{table}[t]
\centering
\resizebox{\columnwidth}{!}{%
\begin{tabular}{ |c|c|c|c| } 
 \hline
 \textbf{Adversarial Defense} & \textbf{FGSM} & \textbf{BIM} & \textbf{DF}\\ 
 \hline
 No-defense & 53.4\%& 32.77\%\%& 44.64\% \\
 JPEG(Q=90) & 53.75\%& 34.39\%&   10.86\% \\ 
 Bit Depth Reduction (4-bit) & 53.76\% & 33.21\% & 27.86\%\\ 
 Bit Depth Reduction (5-bit) &  53.63\% & 33.03\% &  24.86\% \\ 
 Median Filtering (2x2) & 53.22\% & 34.91\%&  28.94\%\\ 
 Median Filtering (3x3) & 53.8\%& 35.08\%& 27.30\% \\
 NL-Means & 53.79\% & 34.24\% &  22.94\%\\
 TVM & 52.53\% & 31.72\%& 30.90\% \\
 Pixel Deflection (W/o RCAM) & 53.46\%& 32.26\%&  27.54\%\\
 Pixel Deflection (W/ RCAM) & 53.5\%& 32.25\%& 25.90\% \\
 \hline
 Saak Transform & \textbf{15.36\%}& \textbf{12.5\%}&  \textbf{4.55\%}\\
 \hline
 \end{tabular}
 }
\caption{Robustness comparison on STL-10: The accuracy on clean images using pre-trained network is 74.86\% and using SAAK transform is 63.5\%.}
\label{table:STL-10}
\end{table}

Table \ref{table:MNIST} shows extensive comparison results of our Saak transform based classification for various attacked MNIST dataset with other state-of-art defense methods. The values across the table indicate the drop in classification accuracy from clean and adversarial attacked images, i.e. drop in classification accuracy $\Delta=C_{clean} -C_{attact}$.  Lower the drop value is, better is the robustness of the classification model. Similarly Tables \ref{table:CIFAR-10} and \ref{table:STL-10} shows the results for CIFAR-10 and STL-10 datasets. 

From the results we can see our approach outperforms other adversarial defense methods. The classification accuracy drop for Saak transform features is very less, thanks for robustness of Saak transform to adversarial perturbations. As previously stated, the images classified with Saak transform are not subjected to any specially-crafted adversarial defense method. The drop obtained for Deepfool attacked images is less when compared to the other attacks. For the MNIST dataset, Saak transform classification accuracy drop is lower than all other defenses for all the three attacks. Also the range of drop is much lesser for MNIST when compared to CIFAR-10 and STL-10, mainly because the attacks are more effective in complex datasets. Even in the complex datasets, the drop in accuracy using Saak features is less and much lower than most of the defenses. The results clearly show that adversarial perturbations can be effectively and efficiently defended using state-of-the-art Saak transform.

\section{Conclusion} \label{sec:Conclusion}
This paper studies the robustness of Saak transform against adversarial attacks without using any additional overhead to remove adversarial noise from images. 
We apply multi-stage Saak transform to build a complete image classification pipeline and carefully design each steps of feature selection, representation and classification to increase adversarial robustness. Extensive experimental evaluations demonstrate the benefits and utilities of Saak transform.

\bibliographystyle{IEEEbib}
\bibliography{bib}

\end{document}